\documentclass[review]{elsarticle}

\usepackage{lineno,hyperref}
\modulolinenumbers[5]

\usepackage{amsmath,epsfig}
\usepackage{balance}
\usepackage{graphicx}
\usepackage{lipsum}
\usepackage{url}
\graphicspath{{Figures/}}
\usepackage{caption}
\usepackage{subfig} 
\usepackage{color}
\usepackage{balance}
\usepackage{float}
\usepackage{multirow}
\usepackage{float}
\usepackage{breqn}
\usepackage{tabularx}

\journal{Healthcare Analytics}

\begin{document}






%
\begin{frontmatter}

\title{A predictive analytics approach for stroke prediction using machine learning and neural networks}

\author[add1,add2]{Soumyabrata~Dev\corref{mycorrespondingauthor}}
\cortext[mycorrespondingauthor]{Corresponding author. Tel.: + 353 1896 1797.}
\ead{soumyabrata.dev@ucd.ie}
\author[add3,add4]{Hewei~Wang}
\ead{hewei.wang@ucdconnect.ie}
\author[add5]{Chidozie~Shamrock~Nwosu}
\ead{x17161916@student.ncirl.ie}
\author[add1]{Nishtha~Jain}
\ead{nishtha.jain@adaptcentre.ie}
\author[add6]{Bharadwaj~Veeravalli}
\ead{elebv@nus.edu.sg}
\author[add7]{Deepu~John}
\ead{Deepu.john@ucd.ie}

\address[add1]{ADAPT SFI Research Centre, Dublin, Ireland}
\address[add2]{School of Computer Science, University College Dublin, Ireland}
\address[add3]{Beijing University of Technology, Beijing, China}
\address[add4]{Beijing-Dublin International College, Beijing, China}
\address[add5]{National College of Ireland, Dublin, Ireland}
\address[add6]{Department of Electrical and Computer Engineering, National University of Singapore, Singapore, Singapore}
\address[add7]{School of Electrical and Electronic Engineering, University College Dublin}

\begin{abstract}
The negative impact of stroke in society has led to concerted efforts to improve the management and diagnosis of stroke. With an increased synergy between technology and medical diagnosis, caregivers create opportunities for better patient management by systematically mining and archiving the patients' medical records. Therefore, it is vital to study the interdependency of these risk factors in patients' health records and understand their relative contribution to stroke prediction. This paper systematically analyzes the various factors in electronic health records for effective stroke prediction. Using various statistical techniques and principal component analysis, we identify the most important factors for stroke prediction. We conclude that age, heart disease, average glucose level, and hypertension are the most important factors for detecting stroke in patients. Furthermore, a perceptron neural network using these four attributes provides the highest accuracy rate and lowest miss rate compared to using all available input features and other benchmarking algorithms. As the dataset is highly imbalanced concerning the occurrence of stroke, we report our results on a balanced dataset created via sub-sampling techniques.
\end{abstract}


\begin{keyword}
predictive analytics ; machine learning ; neural network ; electronic health records ; stroke
\end{keyword}

\end{frontmatter}

\section{Introduction}
\label{sec:intro}
We have witnessed amazing developments in the field of medicine with the aid of
technology~\cite{sivapalan2022annet}. With the advent of annotated dataset of medical records, we can now
use data mining techniques to identify trends in the dataset. Such analysis has
helped the medical practitioners to make an accurate prognosis of any medical
conditions. It has led to an improved healthcare conditions and reduced
treatment costs. The use of data mining techniques in medical records have
great impact on the fields of healthcare and
bio-medicine~\cite{koh2011data,yoo2012data}. This assists the medical
practitioners to identify the onset of disease at an earlier stage. We are
particularly interested in stroke, and to identify the key factors that are
associated with its occurrence. 

Several
studies~\cite{meschia2014guidelines,harmsen2006long,nwosu2019predicting,pathan2020identifying}
have analysed the importance of lifestyle types, medical records of patients on
the probability of the patients to develop stroke. Further, machine learning
models are also now employed to predict the occurrence of
stroke~\cite{jeena2016stroke,hanifa2010stroke}. However, there is no study that
attempts to analyse all the conditions related to patient, and identify the key
factors necessary for stroke prediction. In this paper, we attempt to bridge
this gap by providing a systematic analysis of the various patient records for
the purpose of stroke prediction. Using a publicly available dataset of
$29072$ patients' records, we identify the key factors that are necessary
for stroke prediction. We use principal component analysis (PCA) to transform
the higher dimensional feature space into a lower dimension subspace, and
understand the relative importance of each input attributes. We also benchmark
several popular machine-learning based classification algorithms on the dataset
of patient records. 

The main contributions of this paper are as follows -- (a) we provide a detailed understanding of the various risk factors for stroke prediction. We analyse the various factors present in Electronic Health Record (EHR) records of patients, and identify the most important factors necessary for stroke prediction; (b) we also use dimensionality reduction technique to identify patterns in low-dimension subspace of the feature space; and (c) we benchmark popular machine learning models for stroke prediction in a publicly available dataset. We follow the spirit of reproducible research, and therefore the source code of all simulations used in this paper are available online.\footnote{In the spirit of reproducible research, the code and data to reproduce the results in this manuscript are available online here: \url{https://github.com/Soumyabrata/EHR-features}.}

The structure of the paper is as follows. The remaining part of  Section~\ref{sec:intro} provides an overview of the related work, and describes the dataset used in our study. Section~\ref{sec:analyse-ehr} covers the correlation analysis and feature importance analysis. The results from Principal Component Analysis are explained in  Section~\ref{sec:pca}. The data mining algorithms used for predictive modelling and their performance on the dataset is detailed in  Section~\ref{sec:detection}. Finally, Section~\ref{sec:conc} concludes the paper and discusses future work.

\subsection{Related Work}

Existing works in the literature have investigated various aspects of stroke
prediction. Jeena \textit{et al.}  provides a study of various risk factors to understand
the probability of stroke~\cite{jeena2016stroke}. It used a regression-based
approach to identify the relation between a factor and its corresponding impact
on stroke. In Hanifa and Raja~\cite{hanifa2010stroke}, an improved accuracy for
predicting stroke risk was achieved using radial basis function and polynomial
functions applied in a non-linear support vector classification model. The risk
factors identified in this work were divided into four groups --- demographic,
lifestyle, medical/clinical and functional. Similarly, Luk \textit{et al.}  studied
$878$ Chinese subjects to understand if age has an impact on stroke
rehabilitation outcomes~\cite{luk2006does}. Min \textit{et al.}
in~\cite{min2018development} developed an algorithm for predicting stroke from
potentially modifiable risk factors. Singh and Choudhary
in~\cite{singh2017stroke} have used decision tree algorithm on Cardiovascular
Health Study (CHS) dataset for predicting stroke in patients. A deep learning
model based on a feed-forward multi-layer artificial neural network was also
studied in~\cite{chantamitprediction} to predict stroke. Similar work was
explored in~\cite{khosla2010integrated,hung2019development,teoh2018towards} for
building an intelligent system to predict stroke from patient records. Hung
\textit{et al.}  in~\cite{hung2017comparing} compared deep learning models and machine
learning models for stroke prediction from electronic medical claims database.
In addition to conventional stroke prediction, Li \textit{et al.}
in~\cite{li2016integrated} used machine learning approaches for predicting
ischaemic stroke and thromboembolism in atrial fibrillation.

The results from the various techniques are indicative of the fact that multiple factors can affect the results of any conducted study. These various factors include the way the data was collected, the selected features, the approach used in cleaning the data, imputation of missing values, randomness and standardization of the data will have an impact on the outcome of any study carried. Therefore, it is important for the researchers to identify how the different input factors in an electronic health record are related to each other, and how they impact the final stroke prediction accuracy. 

Studies in related areas~\cite{garcia2016tutorial,yoo2012data} demonstrate that
identifying the important features impacts the final performance of machine
learning framework. It is important for us to identify the perfect combination
of features, instead of using all the available features in the feature space.
As indicated in~\cite{yoo2012data}, redundant attributes and/or totally
irrelevant attributes to a class should be identified and removed before the
use of a classification algorithm. Therefore, it is essential for data mining
practitioners in healthcare to identify how the risk factors captured in
electronic health records are inter-dependent, and how they impact the accuracy
of stroke prediction independently.

\subsection{Electronic Health Records Dataset}

An Electronic Health Record (EHR) also known as Electronic Medical Record
(EMR), is a repository of information for a patient. It is an automated,
computer readable storage of the medical status of a patient that is keyed in
by qualified medical practitioners. The records contain vitals, diagnosis or
medical exam results of a patient. The future of medical diagnosis looks
promising with the optimal use of EHR. The use of EHR increased from 12.5\% to
75.5\% in US Hospitals between 2009 and 2014 as indicated by the statistics
recorded in~\cite{goldstein2017opportunities}. 

For our study, we use a dataset of electronic health records released by McKinsey \& Company as a part of their healthcare hackathon challenge.\footnote{\url{https://datahack.analyticsvidhya.com/contest/mckinsey-analytics-online-hackathon/}.} The dataset is available from Kaggle,\footnote{\url{https://www.kaggle.com/fedesoriano/stroke-prediction-dataset}.} a public data repository for datasets. The dataset contains the EHR records of $29072$ patients. It has a total of $11$ input attributes, and $1$ output feature. The output response is a binary state, indicating if the patient has suffered a stroke or not. The remaining $11$ input features in EHR are: patient identifier, gender ($G$), age ($A$), binary status if the patient is suffering from hypertension ($HT$) or not, binary status if the patient is suffering from heart disease ($HD$) or not, marital status ($M$), occupation type ($W$),  residence (urban/rural) type ($RT$), average glucose level ($AG$), body mass index ($BMI$), and patient's smoking status ($SS$). The dataset is highly unbalanced with respect to the occurrence of stroke events; most of the records in the EHR dataset belong to cases that have not suffered from stroke. The publisher of the dataset has ensured that the ethical requirements related to this data are ensured to the highest standards. In the subsequent discussion of this paper, we will exclude the patient identifier as one of the input feature. We will consider the remaining $10$ input features, and $1$ response variable, in our study and analysis.

\section{Analysing Electronic Health Records}

\label{sec:analyse-ehr}
In this section, we provide an analysis of electronic health records dataset. We perform correlation analysis of the features. We use the entire dataset of EHR records to perform such analysis on the input features of the EHR records. Correlation analysis is useful for feature selection in the following manner: if two features have very high correlation, one of them can be ignored in the prediction of occurrence of stroke as it does not contribute any additional knowledge to the prediction model. Moreover, we evaluate the behaviour of the features individually and in a group to gaze the importance of each individual feature in predicting the occurrence of a stroke. A systematic analysis of the input feature space is an integral part for stroke prediction. It is important to find the optimal and minimal set of predictive features to reduce the computational cost of modelling and efficient archival of EHR records. This paves us the path for clinicians to record \textit{only} those features in the EHR records that are most efficient for stroke prediction.

\subsection{Correlation between features}

\begin{figure}[htb]
  \begin{center}
    \includegraphics[width=0.85\textwidth]{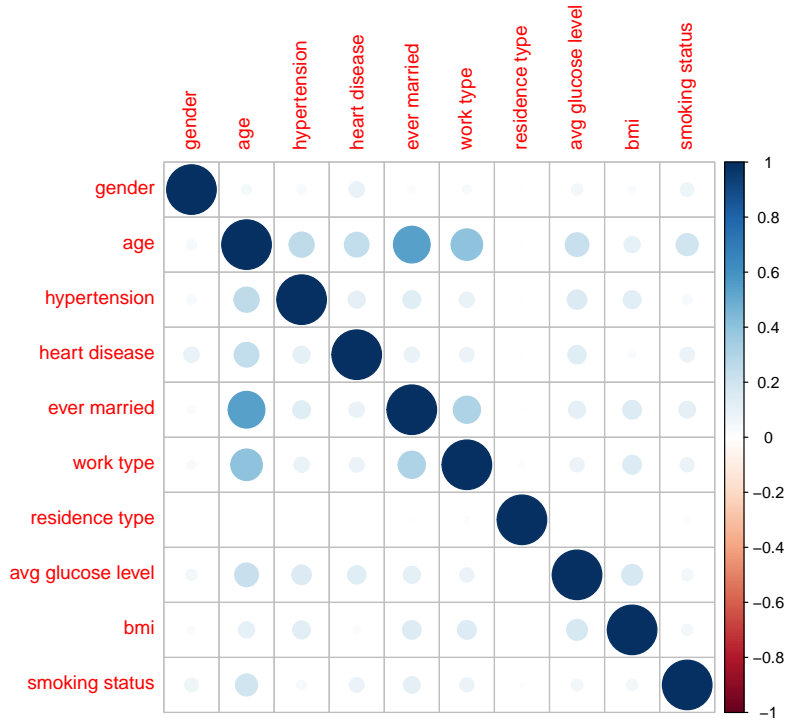}
  \end{center}
  \caption{Correlation matrix for patient attributes in the dataset. These attributes are gender, age, status 0/1 if patient is suffering from hypertension, status 0/1 if patient is suffering from heart disease, marital status, work type, residence type, average glucose level, body mass index and patient's smoking status.}
  \label{fig:corr-matrix}
\end{figure}

We use Pearson's correlation coefficient to generate  Fig.~\ref{fig:corr-matrix}, which shows the correlation between different patient attributes. The strength of the linear relationship between any two features of the patient's electronic health data will be determined by this correlation value. We have used a colourmap in  Fig.~\ref{fig:corr-matrix}, such that the blue colour represents positive correlation, while red is negative. The deeper the colour and larger the circle size, the higher is the correlation between the two patient attributes. 

As is intuitive, the correlation of an attribute with itself is unity. There is a significant correlation between a patient's marital status and their age with 0.5 correlation index. There is also a positive correlation between patient's age and the type of their work with 0.38 correlation index, whether they suffer from hypertension and heart disease or not and their average glucose level. This correlation of patient's age with other attributes seems intuitive, as most ailments occur in an ageing population. The type of residence of patient is not correlated with any other attribute. Patient's type of work has a positive correlation with their marital status with 0.35 correlation index.

In summary, the correlation matrix shown in  Fig.~\ref{fig:corr-matrix} tells us that none of the features are highly correlated with each other. Thus, each feature might have their individual contribution towards stroke prediction. The next two subsections analyse the importance of an individual feature for stroke prediction.

\subsection{Individual Features for stroke prediction}

\label{sec:indi_feat}
\begin{figure}[htb]
  \begin{center}
    \includegraphics[width=0.85\textwidth]{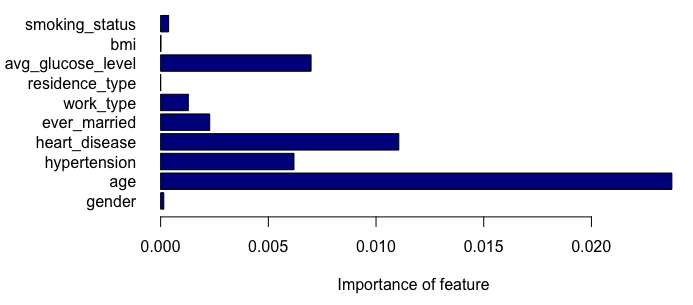}
  \end{center}
  \caption{Importance of patient attributes in predicting the occurrence of stroke with a Linear Vector Quantization (LVQ) model.}
  \label{fig:feat-importance}
\end{figure}

 Fig.~\ref{fig:feat-importance} shows the importance of each patient's attribute in predicting the occurrence of stroke using a Learning Vector Quantization (LVQ) model. The relative importance of a patient's attribute is measured by the increase in the model's prediction error due to that attribute. We use the varImp method from the R \texttt{caret} package\footnote{\url{http://topepo.github.io/caret/index.html}.} to compute this relative feature importance. As  Fig.~\ref{fig:feat-importance} illustrates, patient's age ($A$) is the feature with highest importance in predicting the occurrence of stroke. The other features with high importance are presence of heart disease ($HD$), patient's average glucose level ($AG$) and presence of hypertension.

\begin{figure}[htbp]
  \begin{center}
    \includegraphics[width=0.6\textwidth]{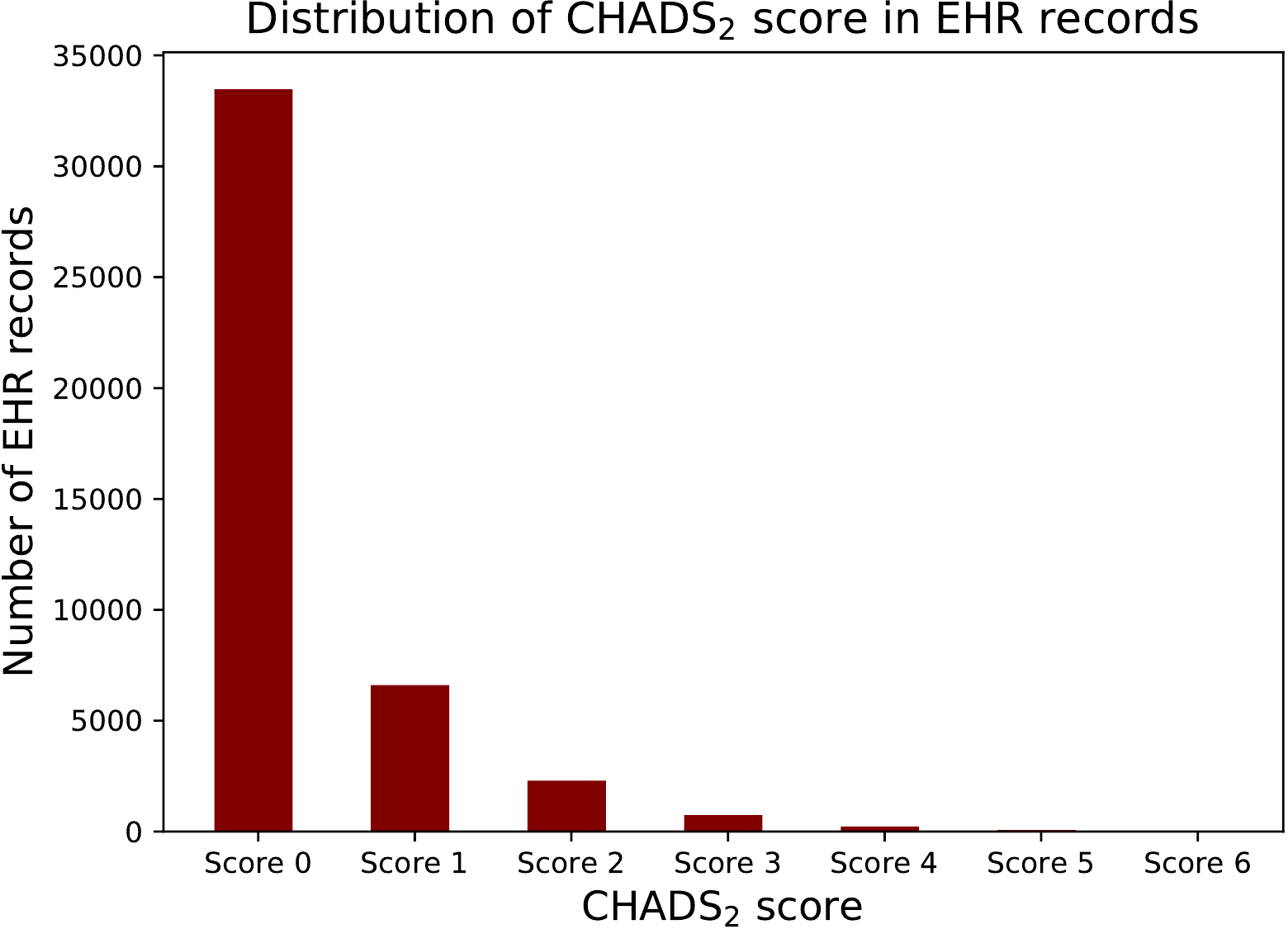}
  \end{center}
  \caption{We show the distribution of CHADS$_2$ score in the EHR records dataset. We observe that most of the CHADS$_2$ score values are low.}
  \label{fig:CHADS1}
\end{figure}

\begin{figure}[htbp]
  \begin{center}
    \includegraphics[width=0.7\textwidth]{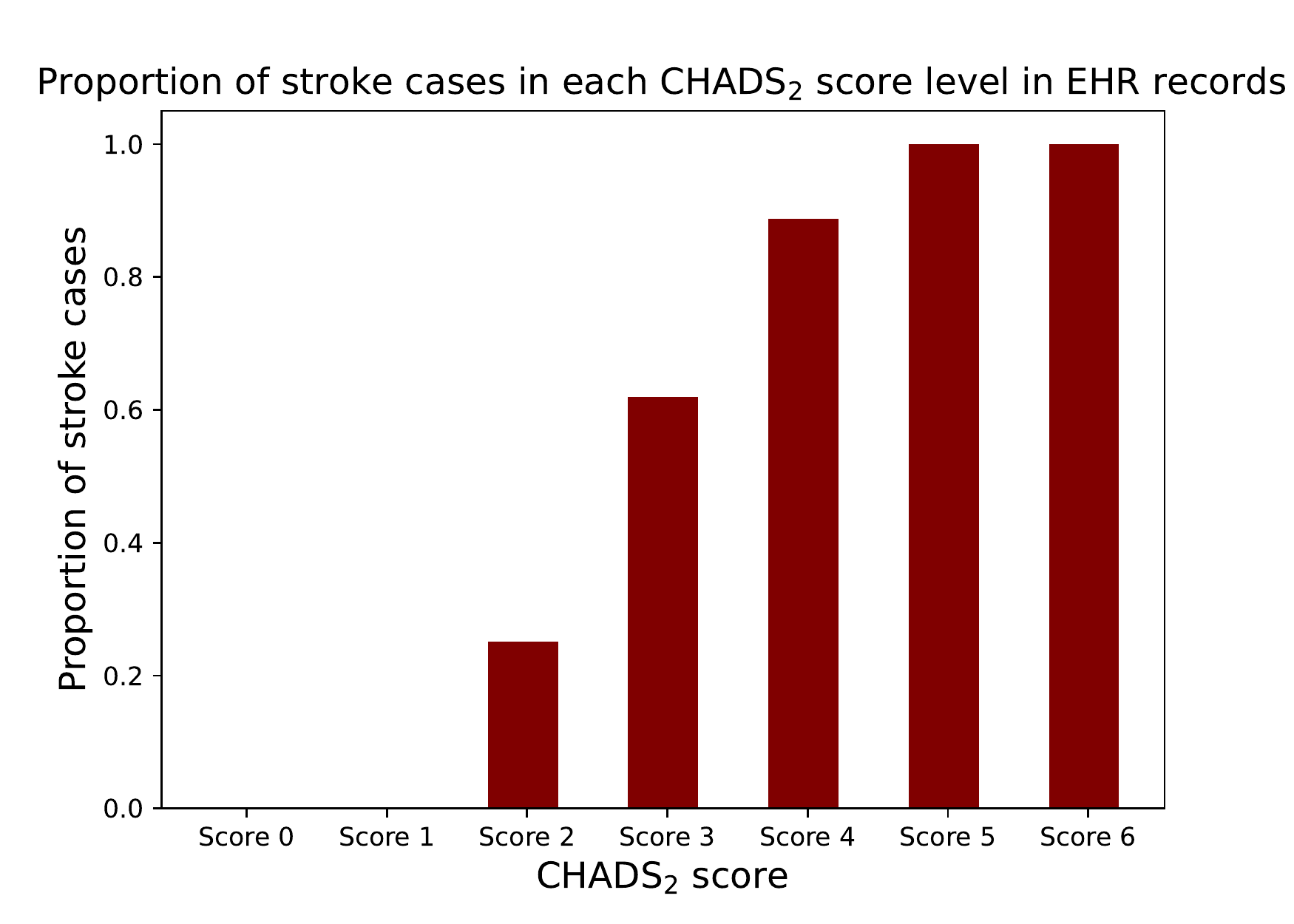}
  \end{center}
  \caption{The proportion of cases with a stroke event as predicted by CHADS$_2$ for each score level respectively}
  \label{fig:CHADS2}
\end{figure}

The analysis described above shows 
 patient's age ($A$) has a comparatively higher importance by itself, yet a combination of different features may improve prediction because they are not correlated with each other. Furthermore, we also compute the CHADS$_2$ score for the EHR records. CHADS2 score is a stroke risk score for non valvular atrial fibrillation, where $C$ represents congestive heart failure or impairment of left ventricular function, $H$ represents whether the patient has hypertension, $A$ stands for age, $D$ stands for diabetes, $S$ stands for stroke, or transient ischaemic attack, history of thromboembolism.  Fig.~\ref{fig:CHADS1} shows the distribution of the CHADS$_2$ score for our dataset. We observe that most of the EHR observations have a low CHADS$_2$ score.  Fig.~\ref{fig:CHADS2} shows the proportion of cases with a stroke event as predicted by CHADS2 for each score level. We observe that the larger the CHADS2 score, the higher is the occurrence of stroke cases. Combined with  Fig.~\ref{fig:CHADS1} and Fig.~\ref{fig:CHADS2}, we find that most people having CHADS$_2$ score of 1 or 2, have a low probability of stroke. We observe that only a small number of people with CHADS$_2$ score greater than 2 have a higher probability of occurrence of stroke.

\subsection{Selection of Optimum Features for stroke prediction}

In the previous section, we used the features one at a time and saw that there are only few features which have higher importance in stroke prediction. In this section, we use all features then subsequently remove one feature at a time or add one feature at a time and further analyse the importance of an individual feature in stroke prediction. We use neural network algorithm to produce the results. We use a perceptron neural network for this experiment. We use the entire dataset of EHR records to perform the feature analysis.

\begin{figure}[htb]
  \begin{center}
    \subfloat[Adding features]{\includegraphics[width=0.65\textwidth]{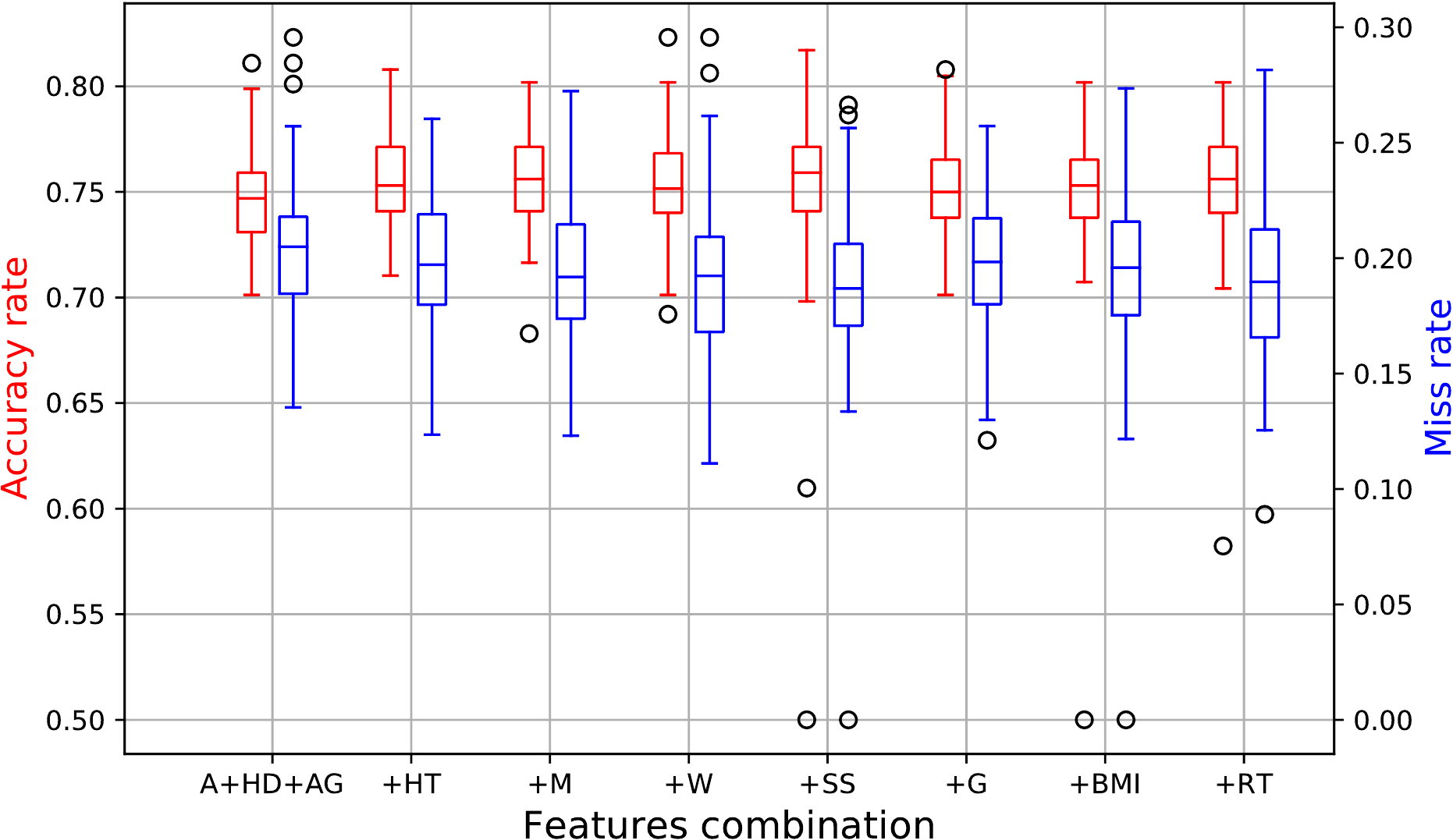}}\\
    \subfloat[Removing features]{\includegraphics[width=0.65\textwidth]{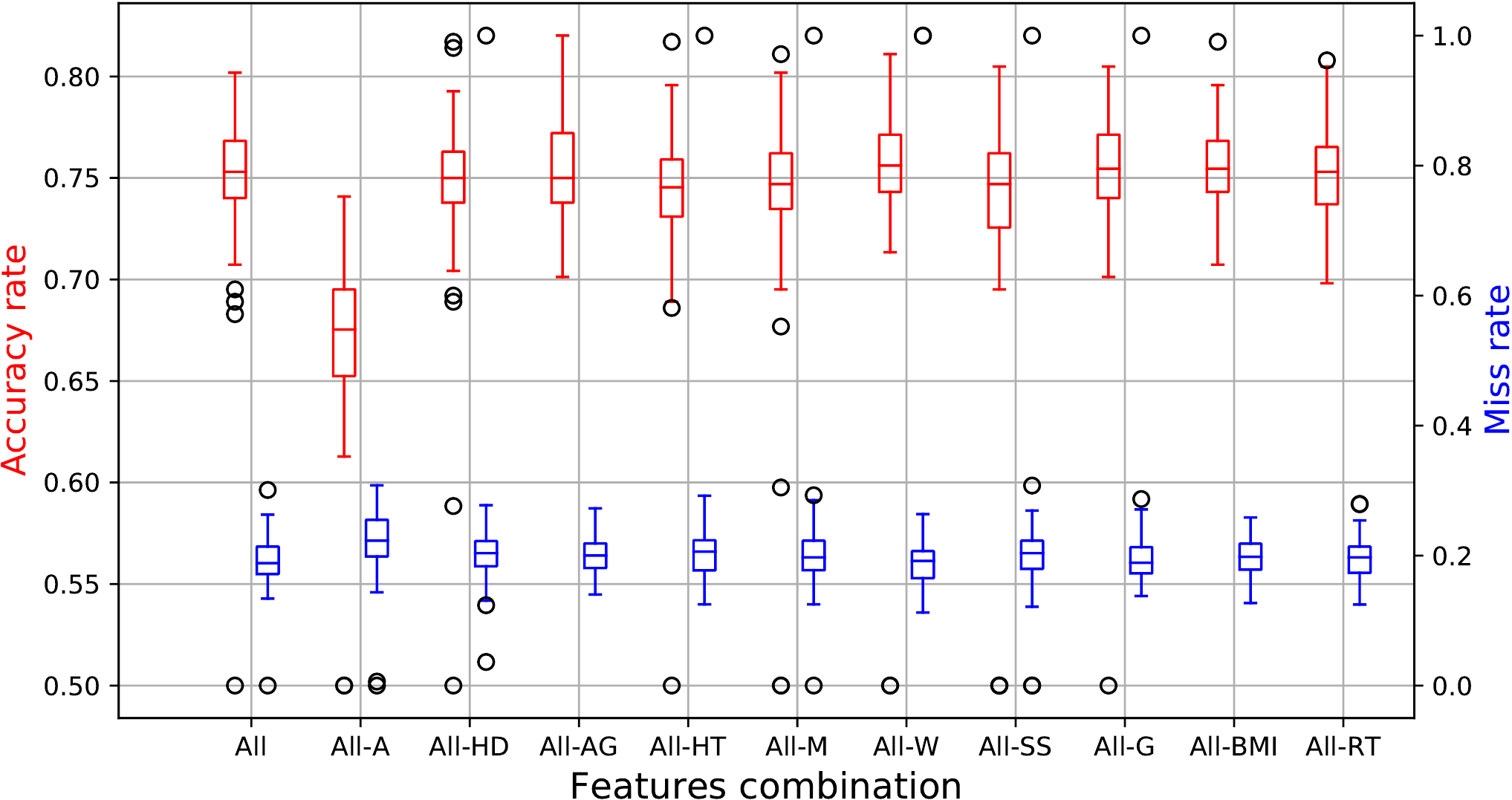}}
  \end{center}
  \caption{We measure the accuracy rate and miss rate of stroke prediction when (a) when features are added one at a time, and (b) individual features are deleted one at a time. The box plots represent the distribution of the metrics, computed from $100$ experiments.}
  \label{fig:feature-add-del}
\end{figure}

 Fig.~\ref{fig:feature-add-del}(a) shows the results for subsequently adding one feature at a time. The first result corresponds to using features $A$, $HD$ and $AG$. These features are chosen as from  Fig.~\ref{fig:feat-importance}, we observe that these three features show higher importance compared to others. When $HT$ is added to this pool, we see that the accuracy rate is slightly improved and miss rate is slightly decreased. Note that $HT$ is the fourth important variable from the analysis of  Fig.~\ref{fig:feat-importance}. Subsequently, when other features are added one by one, the accuracy rate and miss rate show very slight variation. The accuracy and miss rate are almost same for all the remaining configurations. Therefore, it suggests that the four features: $A$, $HD$, $AG$ and $HT$ can be optimum features as there is no improvement offered by addition of other features.

 Fig.~\ref{fig:feature-add-del}(b) shows the results for deleting one feature at a time from the pool of all features. Here we can observe that the accuracy rate is significantly affected when the feature $A$ is removed from the pool. This is inline with our earlier discussion which showed that $A$ has highest score amongst all features (ref to  Fig.~\ref{fig:feat-importance}). There is no much significant changes when other features are removed from the pool.

\section{Principal Component Analysis}

\label{sec:pca}

In this section, we analyse the variance in the dataset using Principal Component Analysis (PCA). In this multivariate analysis, the dataset is transformed into a set of values of linearly uncorrelated variables called principal components such that maximum variance is extracted from the variables. These principal components act as summaries of the features of the dataset. These new basis functions do not have a physical interpretation. However, these new basis functions are linear combinations of the original feature vectors. In this work, we do not restrict ourselves on the feature analysis using traditional feature elimination techniques. However, we use dimensionality reduction technique to transform the high-dimensional feature space onto 2-dimensional feature space to understand the inter-relation amongst the feature space. PCA can be used to reduce the feature space for predictive modelling if the first few components capture most of the variance in the data. We analyse the $10$ dimensional patient attribute in the lower dimension subspace using PCA. We provide a brief primer on principal component analysis and mathematically formulate our problem statement.

Let us suppose that $\mathbf{X}$ is the variable matrix of dimension $m \times n$. In this case, $m$ indicates the total number of input attributes in EHR, and $n$ is the total number of patient records in the dataset. Therefore, $m=10$ and $n=29072$ in this analysis for stroke prediction. We vectorize the individual features $f_{1-10}$ from the matrix $\mathbf{X}$, into corresponding $\widetilde{\mathbf{v}}_j \in \mathbf{R}^{mn \times 1}$ where $j=1,2,\ldots,10$. Finally, the $\widetilde{\mathbf{v}}_j$ features are stacked together to create the matrix $\hat{\mathbf{X}} \in \mathbf{R}^{mn \times 10}$:

\begin{equation}
\label{eq:eq1}
\hat{\mathbf{X}}=[\widetilde{\mathbf{v}}_1, \widetilde{\mathbf{v}}_2,\widetilde{\mathbf{v}}_3,\ldots,\widetilde{\mathbf{v}}_{10}].
\end{equation}

We perform the PCA on the normalized matrix of $\hat{\mathbf{X}}$, that is normalized using the corresponding means $\bar{v_{j}}$ and standard deviations $\sigma_{v_{j}}$ of the individual features. The normalized matrix $\ddot{\mathbf{X}}$ is represented as:
\begin{equation}
\label{eq:eq3}
\ddot{\mathbf{X}}= \left[\frac{\widetilde{\mathbf{v}_{1}}-\bar{v_{1}}}{\sigma_{v_{1}}}, \frac{\widetilde{\mathbf{v}_{2}}-\bar{v_{2}}}{\sigma_{v_{2}}},\ldots,\frac{\widetilde{\mathbf{v}_{j}}-\bar{v_{j}}}{\sigma_{v_{j}}},\ldots,\frac{\widetilde{\mathbf{v}_{10}}-\bar{v_{10}}}{\sigma_{v_{10}}}\right].
\end{equation}

We interpret how the results from PCA are related to predictive variables or
features represented by patient attributes and the individual observations
represented by the medical health records. We study the relation between the
first two principal components and the individual input variables. We also
study the importance of the first two principal components for a given
observation. We use the guide by Abdi and Williams~\cite{abdi2010principal} for
this study.

\subsection{Variance explained by principal components}

\begin{figure}[htb]
  \begin{center}
    \subfloat[]{\includegraphics[width=0.45\textwidth]{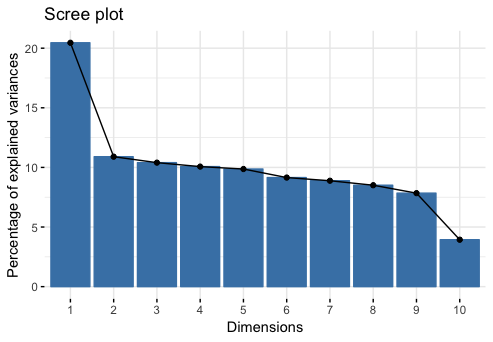}}
    \subfloat[]{\includegraphics[width=0.45\textwidth]{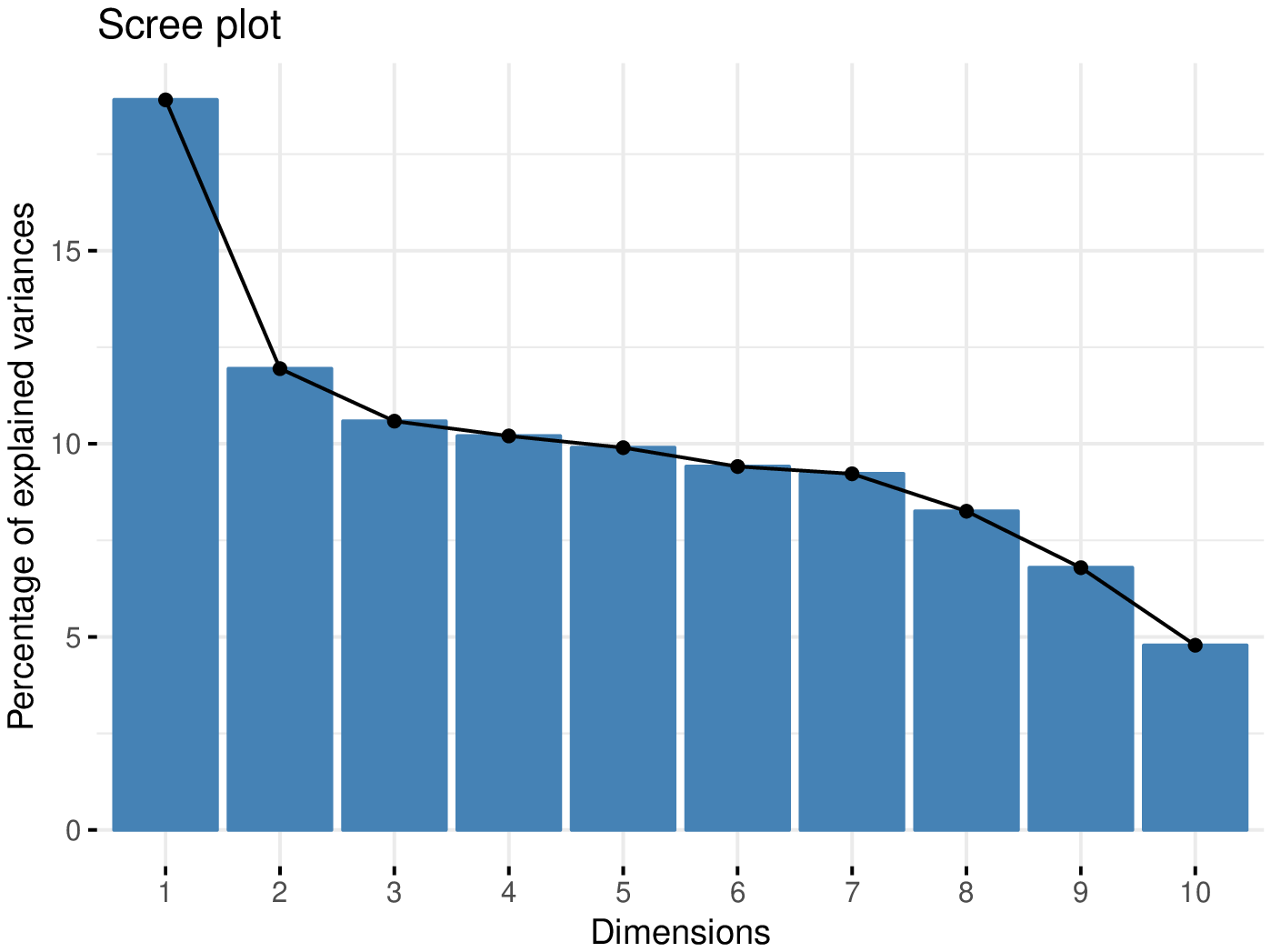}}
  \end{center}
  \caption{Percentage of variance explained by different principal components. We show the scree plot for (a) original, (b) balanced datasets.}
  \label{fig:scree}
\end{figure}

 A scree plot is used to select the components which explain most variability in the dataset, generally $80$\% or more variance.  Fig.~\ref{fig:scree}(a) shows the percentage of variance in the dataset explained by the different principal components in the original dataset. Here, we can observe that the variance explained by different principal components are very low. Out of $10$ principal components, $8$ are needed to explain variance of 88.2\%. Balanced dataset means that the dataset is balanced with respect to the stroke labels using random sampling. The original dataset is unbalanced because there are more samples possessing negative stroke labels, as compared to positive label stroke samples. We make it balanced by considering all the positive stroke samples, and then randomly picking equal number of negative stroke samples from the rest. This will make a balanced dataset with equal number of positive and negative stroke samples.

All principal components are orthogonal to each other and hence  uncorrelated.   Therefore, each individual PC can be useful to explain a unique phenomenon. The distributed variance in  Fig.~\ref{fig:scree} indicates that the different principal components are explaining different underlying phenomenon. These phenomenon can be analysed based on the variable loadings. Variable loadings are the contribution of different variables to an individual principal component. We have included the scree plot for the balanced dataset as well in  Fig.~\ref{fig:scree}(b). This is useful for researchers to understand the impact of unbalanced nature of the dataset on the subspace representation. 
Table~\ref{table:pc-loadings} shows the contribution of each variable towards first two principal components. The sum of the squares of all loadings for an individual principal component equals unity. Therefore, if the variable loading crosses a threshold value of $\sqrt{1/10}=0.31$, it indicates that the variable has a strong contribution towards the principal component. In 
Table~\ref{table:pc-loadings}, the variables that cross the threshold are in bold. Here we observe that variables like $A$, $AG$, $HT$ and $M$ have strong contribution towards the first principal component. The variable $HD$ also shows significant contribution towards it. From earlier discussions, we saw that these features actually are important from stroke prediction point of view. Therefore, it indicates that the first principal component (which has stronger loadings from these variables) might be useful in predicting the stroke.

\begin{table}[htb]
\centering
\normalsize 
\begin{tabular}{l|cc}
\textbf{Features} & $PC_1$ & $PC_2$ \\
\hline 
gender & 0.092 & \textbf{0.516} \\
age & \textbf{0.571} & 0.230 \\
hypertension & \textbf{0.331} & 0.232 \\
heart\_disease & 0.290 & 0.261 \\
ever\_married & \textbf{0.475} & \textbf{0.322} \\
work\_type & 0.236 & \textbf{0.442} \\
residence\_type & 0.001 & 0.003 \\
avg\_glucose\_level & \textbf{0.326} & \textbf{0.403} \\
bmi & 0.242 & 0.293 \\
smoking\_status & 0.152 & 0.090
\end{tabular}
\caption{We check the contributions of the different features in the first and second principal components. We report the absolute values of the different loading factors.}
\label{table:pc-loadings}
\end{table}

In the following sections, we will assess the role of these principal components in stroke prediction.

\subsection{Relation between principal components with patient attributes}

\begin{figure}[htb]
  \begin{center}
    \subfloat[]{\includegraphics[width=0.48\textwidth]{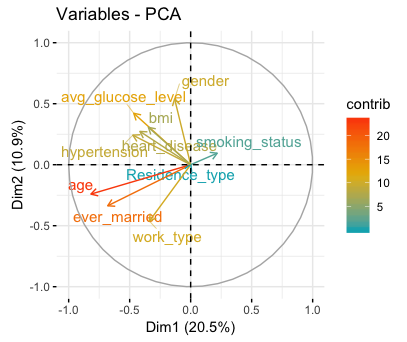}}
    \subfloat[]{\includegraphics[width=0.52\textwidth]{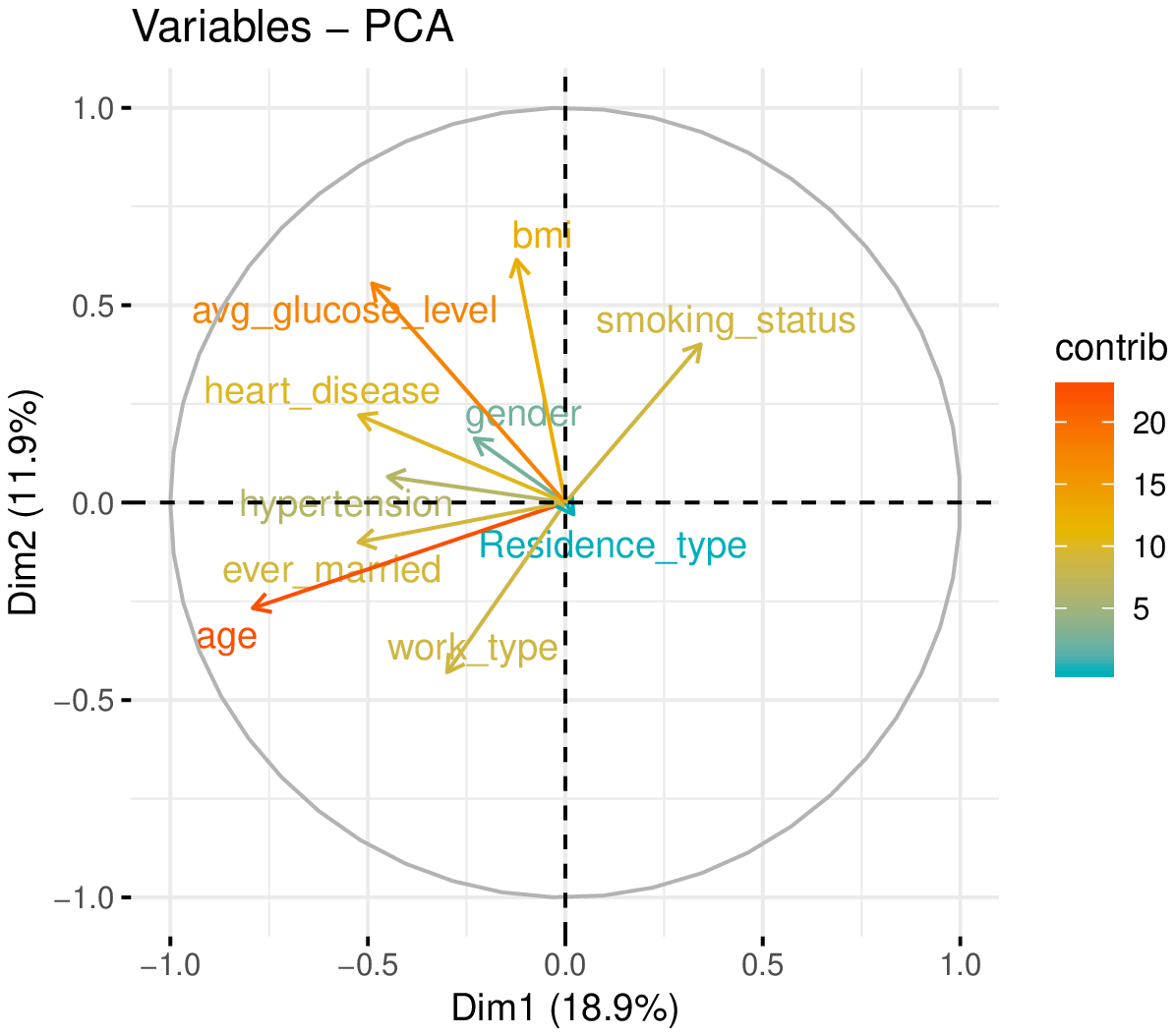}}
  \end{center}
  \caption{Biplot representation of the input attributes on the first two principal components. We show the biplot for (a) original, (b) balanced datasets.}
  \label{fig:biplot}
\end{figure}

 Fig.~\ref{fig:biplot} describes the biplot representation. It shows how the different input attributes are correlated with other, and also depend on the first and second principal components. The $x$-axis and $y$-axis indicate the first and second principal components respectively. The each vectors represent an input attribute, and its length indicate its importance. We observe that the average glucose level and the heart disease are correlated to each other. The age has the biggest contribution in the first two principal component. We also observe that the orientation of the different feature vectors in the two-dimensional feature space is the same as the unbalanced dataset.

 Fig.~\ref{fig:biplot}(a) illustrates that the contribution of patient's residence type is minimum to the two principal components. We also observe that age and status of marriage are correlated with each other, and have a high contribution to the first principal component. The smoking status of a patient and its average glucose level are orthogonal to each other, indicating that they provide different information to the feature space. However, smoking status and age point opposite to each other, indicating that they provide similar information, but have a diverging characteristics. We also compute the biplot for the EHR records on the balanced dataset. We show the corresponding biplot in  Fig.~\ref{fig:biplot}(b). We observe that the average glucose level and the heart disease are correlated to each other. The age has the biggest contribution in the first two principal component. We also observe that the orientation of the different feature vectors in the two-dimensional feature space is the same as the unbalanced dataset.

\subsection{Relation of Principal Components with individual records}

\begin{figure}[htb]
  \begin{center}
    \subfloat[]{\includegraphics[width=0.45\textwidth]{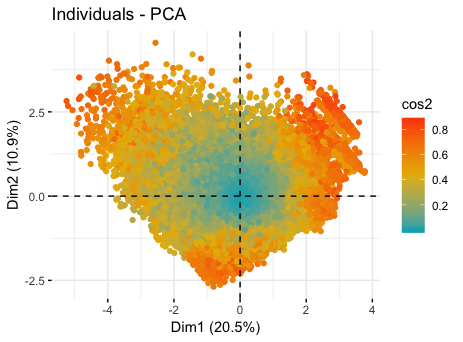}}\quad\quad
    \subfloat[]{\includegraphics[width=0.45\textwidth]{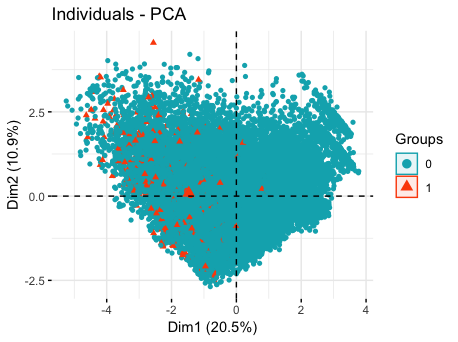}}\\
    \subfloat[]{\includegraphics[width=0.5\textwidth]{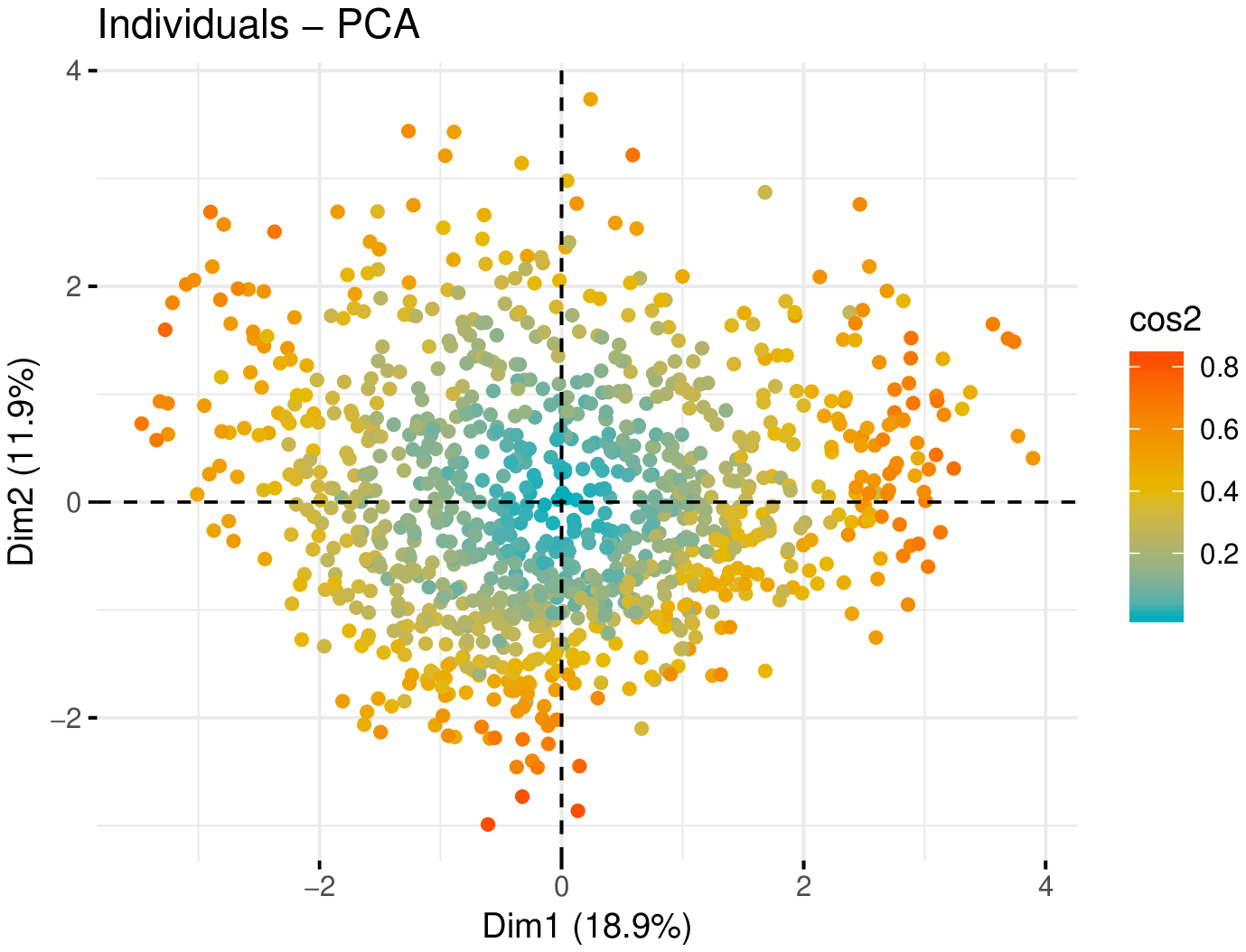}}
    \subfloat[]{\includegraphics[width=0.5\textwidth]{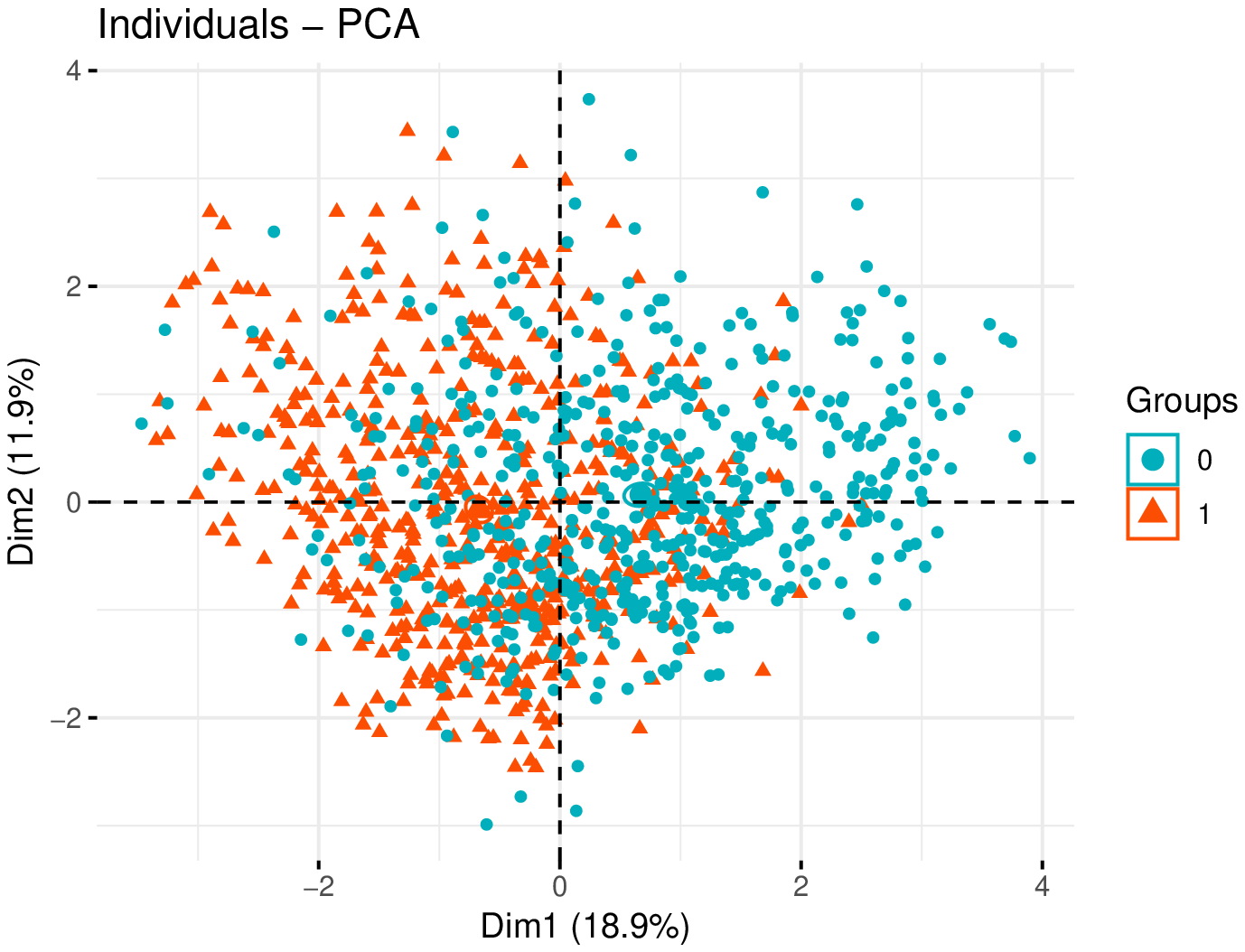}}
  \end{center}
  \caption{Subspace representation of the different patient records in reference to the first two principal components. The observations are colour coded based on (a) \emph{cos2} measure indicating the importance of principal components on the observation; and (b) status of stroke. We also use balanced dataset and observe the sub-space representation in (c) and (d). }
  \label{fig:pca-individual}
\end{figure}

Finally, we also check the relation of individual patient record observations on the first two principal components. In  Fig.~\ref{fig:pca-individual}, each dot represents a patient record observation.

Fig.~\ref{fig:pca-individual}(a) shows the importance of principal components onto each of the records. This is generally represented by the \emph{cos2} measure, indicating the squared distance from the origin. This indicates that observations that possess a high \emph{cos2} value can be represented by the principal components, as opposed to observations with a low \emph{cos2} value. We observe that a significant number of points are clustered near the origin, which cannot be represented completely by the principal components. Therefore, the principal component can indicate only a part of the entire information in the feature space. 

Fig.~\ref{fig:pca-individual}(b) colour codes the patient records based on the status of the stroke. As the dataset is highly unbalanced, we observe that most of the observations are colour coded with negative status of stroke. We observe a few observations with positive status of stroke (observations with $1$ label). However, these positive stroke observations are not located in clusters. This indicates that higher-dimensional features are necessary to separate the observations. 

We also compute the subspace representation of the different features for the balanced EHR dataset. We show it in  Fig.~\ref{fig:pca-individual}(c) and (d). We observe that the observations with \textit{stroke} and \textit{no stroke} are scattered throughout the two principal axes. The observations with similar labels are not clustered together. Therefore, the features of the EHR records are important for efficient stroke prediction.

\subsection{Discussion}

This section discussed how principal component analysis can assist in a clear understanding of the original feature space of patient records. We showed that the first two principal components can cumulatively capture only 31.4\% of the total variance in the input feature space. More so, first eight components can explain only about 88\% of the total variance. Moreover, we studied the contribution of different patient attributes to the first two principal components. We could see that the two components do not represent the health records data perfectly. We also looked at the contribution of the first two principal components in the representation of individual health records. We could see that some health records can be represented by the two components, but some cannot. Thus, all principal components are needed to have a good representation of the variance in medical records. We cannot get a significant reduction of feature space for predictive modelling without a significant loss of variance in the data. Hence, we use all the principal components for predictive modelling of stroke occurrence.

In the next section, we compare the state of art machine learning classification techniques for predicting the occurrence of stroke in a patient's medical record. As discussed, we use ten patient attributes as input features to the models.

\section{Stroke Prediction}

\label{sec:detection}
We provide a detailed analysis of various benchmarking algorithms in stroke prediction in this section. We benchmark three popular classification approaches --- neural network (NN), decision tree (DT) and random forest (RF) for the purpose of stroke prediction from patient attributes. The decision tree model is one of the popular binary classification algorithm. This method involves building a tree-like decision process with several condition tests, and then applying the tree to the medical record dataset. Each node in this tree represents a test, and the branches correspond to the outcome of the test. The leaf nodes finally represent the class labels. The pruning ability of such algorithm makes it flexible and accurate, which is required in medical diagnosis. We also benchmark the dataset on random forest approach. The flexibility and ease of use of the random forest algorithm coupled with its consistency in producing good results, even with minimal tuning of the hyper-parameters makes this algorithm valuable in this application. The possibilities of over-fitting are limited by the number of trees existent in the forest. Moreover, random forest can also provide adequate indicators on the way it assigns significance to each of these input variables. We also benchmark the performance of a 2-layer shallow neural network. Artificial neural networks are quite popular these days, and they offer competitive results. We implement the feed-forward multi-layer perceptron model using the \texttt{nnet} R package.

\subsection{Benchmarking using all features}

Our dataset contains a total of $29072$ medical records. Out of this, only $548$ records belong to patients with stroke condition, and the remaining $28524$ records have no stroke condition. This is a highly unbalanced dataset. This creates problem in using this data directly for training any machine-learning models. Therefore, we use random downsampling technique to reduce the adverse impact of the unbalanced nature of the dataset. We refer the $548$ records as the minority class, and the remaining $28524$ records with no stroke condition as the majority class. Subsequently, we create a dataset of $1096$ observations, that consists of  $548$ minority samples and $548$ majority samples. This balanced dataset is created by considering all the $548$ minority samples, and the remaining $548$ majority samples are selected randomly from the $28524$ patient records. All the three machine learning models are trained on this balanced dataset of $1096$ observations.

\begin{table*}[htb]
\centering
\scriptsize  
\begin{tabular}{l|r|cccccc}
\textbf{Features} & \textbf{Way} & \textbf{Precision} & \textbf{Recall} & \textbf{F-score} & \textbf{Accuracy} & \textbf{Miss rate} & \textbf{Fall-out rate}\\
\hline 
\multirow{3}{*}{\begin{tabular}[c]{@{}l@{}}Original Features\\(All)\end{tabular}}& DT & 0.75 & 0.74 & 0.74 & 0.74 & 0.17 & 0.24\\
& RF & 0.74 & 0.73 & 0.73 & 0.74 & 0.18 & 0.25\\
& NN & 0.80 & 0.74 & 0.77 & 0.77 & 0.16 & 0.18\\
& CNN & 0.74 & 0.72 & 0.73 & 0.74 & 0.17 & 0.24\\
& SVM & 0.67 & 0.68 & 0.68 & 0.68 & 0.23 & 0.32\\
& LASSO & 0.78 & 0.72 & 0.75 & 0.76 & 0.19 & 0.20\\
& ElasticNet & 0.79 & 0.71 & 0.75 & 0.76 & 0.19 & 0.19\\
\hline 
\multirow{3}{*}{\begin{tabular}[c]{@{}l@{}}Original Features\\(A+HD+AG+HT)\end{tabular}}& DT & 0.78 & 0.71 & 0.74 & 0.75 & 0.20 & 0.21 \\
& RF & 0.76 & 0.74 & 0.75 & 0.75 & 0.18 & 0.24 \\
& NN & 0.78 & 0.71 & 0.74 & 0.75 & 0.19 & 0.20 \\
\hline 
\multirow{3}{*}{\begin{tabular}[c]{@{}l@{}}PCA Features\\(PC1 and PC2)\end{tabular}}& DT & 0.78 & 0.65 & 0.71 & 0.73 & 0.24 & 0.19\\
& RF & 0.71 & 0.68 & 0.69 & 0.69 & 0.23 & 0.28\\
& NN & 0.77 & 0.67 & 0.72 & 0.74 & 0.22 & 0.20\\
\hline 
\multirow{3}{*}{\begin{tabular}[c]{@{}l@{}}PCA Features\\(PC1 till PC8)\end{tabular}}& DT & 0.75 & 0.68 & 0.72 & 0.73 & 0.21 & 0.23\\
& RF & 0.73 & 0.69 & 0.71 & 0.72 & 0.21 & 0.25\\
& NN & 0.80 & 0.68 & 0.73 & 0.75 & 0.21 & 0.17
\end{tabular}
\caption{Performance evaluation of neural network, decision tree and random forest on our dataset of electronic medical records. We compare their performance for three different cases -- (a) using all the original features, (b) using the PCA-transformed data of the first two principal components, and (c) using the PCA-transformed data of the first eight components. We choose $8$ components, as $8$ components are necessary to cumulatively contain more than $80\%$ of the explained variance. We report the average value of precision, recall, F-score, accuracy, miss rate and fall-out rate, based on $100$ experiments.}
\label{table:compare}
\end{table*}

In our experiment, another deep learning approach, the convolutional neural network (CNN) is implemented for the prediction of stroke. In our configuration, the number of hidden layers is four while the first two layers are convolutional layers and the last two layers are linear layers, the hyperparameters of the CNN model is given in 
Table~\ref{tab3-hyperparameters}. We use the same train and test split for CNN training and testing procedure, the ten inputs features are reshaped into 1 * 2 * 5 for inputs. 

We also calculate accuracy variance for the benchmarking methods. Table~\ref{variance} shows the experiments of accuracy variance for NN, SVM, LASSO and ElasticNet.

\begin{table}[]
\centering
\scriptsize
\resizebox{\textwidth}{!}{%
\begin{tabular}{l|r|cc}

\textbf{Features}                                                                   & \textbf{Way} & \textbf{Precision} & \textbf{Accuracy Variance} \\ 
\hline
\multirow{4}{*}{\begin{tabular}[c]{@{}l@{}}Original Features \\ (All)\end{tabular}} & NN           & 0.80               & 0.000377                                        \\ 
                                                                                    & SVM          & 0.67               & 0.000470                                        \\  
                                                                                    & LASSO        & 0.78               & 0.000380                                        \\  
                                                                                    & ElasticNet   & 0.79               & 0.000467                                        \\ 
\end{tabular}%
}
\caption{Accuracy variance for NN, SVM, LASSO and ElasticNet}
\label{variance}
\end{table}

\begin{table}[]
\resizebox{\textwidth}{!}{%
\begin{tabular}{l|c|c|c}
\textbf{Layers} & \multicolumn{1}{l|}{\textbf{In channels / Out channels}} & \multicolumn{1}{l|}{\textbf{Kernel, Stride, Padding}} & \textbf{Activation functions} \\ \hline
Conv1           & 1/16                                                     & 3, 1, 1                                               & ReLu                          \\ \hline
Conv2           & 16/8                                                     & 2, 1, 0                                               & ReLu                          \\ \hline
\textbf{Layers} & \multicolumn{1}{l|}{\textbf{In features / Out features}} & \multicolumn{1}{l|}{\textbf{Kernel, Stride, Padding}} & \textbf{Activation functions} \\ \hline
Linear1         & 32/16                                                    & -                                                     & ReLu                          \\ \hline
Linear2         & 16/1                                                     & -                                                     & Sigmoid                       \\ 
\end{tabular}%
}
\caption{Hyperparameters of the CNN model}
\label{tab3-hyperparameters}
\end{table}

\subsection{Benchmarking using top four features}

 Here we present results for stroke prediction when all the features are used and when only $4$ features ($A$, $HD$, $AG$ and $HT$) are used. These features are selected based on our earlier discussions. In addition to the features, we also show results for stroke prediction when principal components are used as the input. Since we observed that almost $8$ principal components are needed to explain a variance of greater than $80$\%, we present results for both the cases when only first $2$ principal components are used and when all the components are used. 
Table~\ref{table:compare} shows the evaluation metrics for all the different configurations. In order to remove sampling bias, we perform $100$ random downsampling experiments. The ratio of the number of training observations and testing observations is $70$: $30$. 
Table~\ref{table:compare} reports the average values for all the approaches. 

We observe that, among different machine learning approaches,\footnote{We do not benchmark our approaches against the McKinsey Kaggle challenge winner, as the model code is not publicly available.} neural network works with better accuracy for different feature combinations. When we compare the neural network results for cases when all features are used and when only $4$ features are used, we do not observe a significant improvement of all features over $4$ features. Therefore, we can get a good stroke prediction accuracy of up to $78$\% with a low miss rate of $19$\% by using only $4$ features ($A$, $HD$, $AG$ and $HT$). This result might not be sufficient to guide treatment and prevention measures on an individual level but it will assist in supporting allocation and resources on a population and/or cohort level. 

Here, for the case when the principal components are used as inputs, we observe that there is only slight improvement in accuracy of neural network, when all components are used compared to the first two components only. This is inline with our earlier discussion, where we illustrated that the first principal component can be important from the stroke prediction point of view. The variables that contributed most to the first component have shown good stroke prediction possibility. Therefore, use of only first two components have similar results compared to the case when all components are used.

When we compare the principal component results to the case when actual features are used, it can be observed that the accuracy of neural networks for both cases are comparable. However, if we look at the miss rate, the miss rates are higher for the case of principal components. The miss rate is an important evaluation metric as we would want to be able to detect all the strokes without a fail. We expect miss rate to be as low as possible and a slight degradation in the miss rate value is important for us for further consideration. Therefore, our analysis suggests that the best possible results for stroke prediction can be achieved by using neural network with $4$ important features ($A$, $HD$, $AG$ and $HT$) as input.

\begin{figure}[htb]
  \begin{center}
    \includegraphics[width=0.85\textwidth]{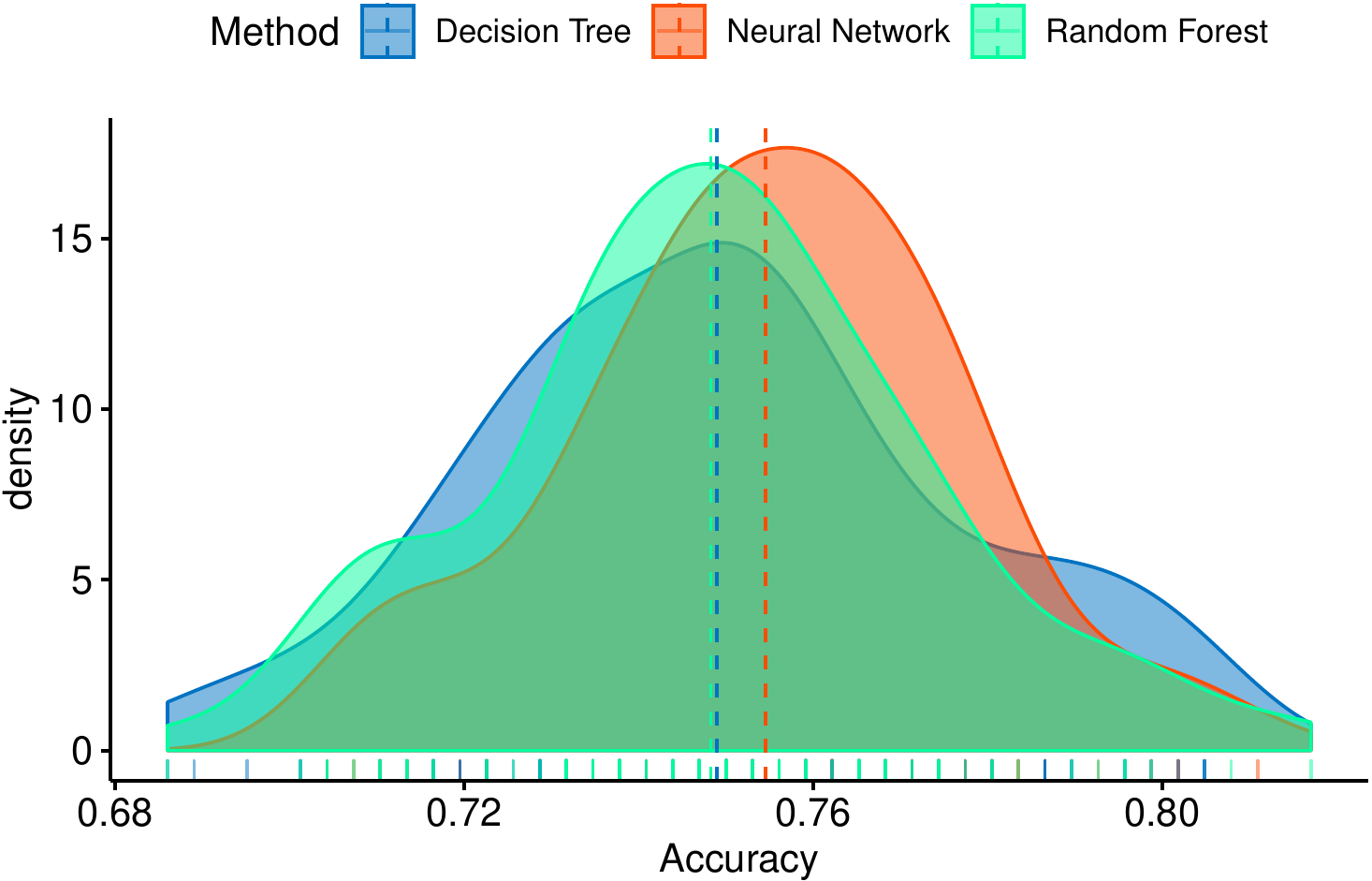}
  \end{center}
  \caption{Histogram distribution of classification accuracies obtained from the $100$ experiments using top 4 features -- age, heart disease, average glucose level, hypertension for the benchmarking algorithms.}
  \label{fig:accurate-dist}
\end{figure}

Finally, we illustrate the distribution of the accuracy values, by using the top 4 features --- age, heart disease, average glucose level, hypertension from the dataset. We perform the experiments $100$ times to remove any sampling bias in the training and testing sets. Fig.~\ref{fig:accurate-dist} illustrates this. We observe that most of them have similar performance, with their mean overlapping around similar values. We also compute the variance of the accuracies of the benchmarking methods for the $100$ random sub-sampling observations. The variance of decision tree, neural network and random forest are $0.00073$, $0.00049$, and $0.00061$ respectively. This indicates that there is no sampling bias involved in the benchmarking results.

\section{Conclusion and Future Work}

\label{sec:conc}

In this paper, we presented a detailed analysis of patients' attributes in electronic health record for stroke prediction. We systematically analysed different features. We performed feature correlation analysis and a step wise analysis for choosing an optimum set of features. We found that the different features are not well-correlated and a combination of only $4$ features ($A$, $HD$, $HT$ and $AG$) might have good contribution towards stroke prediction. Additionally, we performed principal component analysis. The analysis showed that almost all principal components are needed to explain a higher variance. The variable loadings however showed that the first principal component which has the highest variance might explain the underlying phenomenon of stroke prediction. Finally, three machine learning algorithms were implemented on a set of different features and principal components configurations. We found that neural network works the best with a feature combination of $A$, $HD$, $HT$ and $AG$. The accuracy and miss rate for this combination are $78$\% and $19$\% respectively.

We have seen promising results from using just $4$ features. The
accuracy of the perceptron model cannot be improved further for primarily two
reasons: lack of additional discriminatory feature set; and lack of additional
dataset. We observed that most of the existing features in the EHR dataset are
highly correlated to each other, and therefore do not add any additional
information to the original feature space. Furthermore, a larger dataset will
enable us to train our deep neural networks more efficiently. We plan to
collect institutional data in our planned future work. The systematic analysis
of the different features in the electronic health records will assist the
clinicians in effective archival of the records. Instead of recording and
storing all the features, the data management team can archive \textit{only}
those features that are essential for stroke prediction. Thus, in future, we
plan to integrate the electronic records dataset with background knowledge on
different diseases and drugs using Semantic Web
technologies~\cite{tilahun2014design,orlandi2019interlinking}.  Knowledge graph
technologies~\cite{wu2021uplifting,orlandi2019interlinking} can be used in
order to publish the electronic health records in an interoperable manner to
the research community. The added background knowledge from other datasets can
also possibly improve the accuracy of stroke prediction models as well. We
intend to collect our institutional dataset for further benchmarking of these
machine learning methods for stroke prediction. We also plan to perform
external validation of our proposed method, as a part of our upcoming planned
work.

\section{Acknowledgement}
This research was conducted with the financial support of Science Foundation Ireland under Grant Agreement No.\ 13/RC/2106\_P2 at the ADAPT SFI Research Centre at University College Dublin. ADAPT, the SFI Research Centre for AI-Driven Digital Content Technology, is funded by Science Foundation Ireland through the SFI Research Centres Programme.


\begin{thebibliography}{10}
\expandafter\ifx\csname url\endcsname\relax
  \def\url#1{\texttt{#1}}\fi
\expandafter\ifx\csname urlprefix\endcsname\relax\def\urlprefix{URL }\fi
\expandafter\ifx\csname href\endcsname\relax
  \def\href#1#2{#2} \def\path#1{#1}\fi

\bibitem{sivapalan2022annet}
G. Sivapalan, K. Nundy, S. Dev, B. Cardiff, D. John, {ANNet}: A Lightweight Neural Network for {ECG} Anomaly Detection in {IoT} Edge Sensors, IEEE Transactions on Biomedical Circuits and Systems (2022).

\bibitem{koh2011data}
H.~C. Koh, G.~Tan, et~al., Data mining applications in healthcare, Journal of
  healthcare information management 19~(2) (2011) 65.

\bibitem{yoo2012data}
I.~Yoo, P.~Alafaireet, M.~Marinov, K.~Pena-Hernandez, R.~Gopidi, J.-F. Chang,
  L.~Hua, Data mining in healthcare and biomedicine: a survey of the
  literature, Journal of medical systems 36~(4) (2012) 2431--2448.

\bibitem{meschia2014guidelines}
J.~F. Meschia, C.~Bushnell, B.~B-A., L.~T. Braun, D.~M. Bravata, S.~Chaturvedi,
  M.~A. Creager, R.~H. Eckel, M.~S. Elkind, M.~Fornage, et~al., Guidelines for
  the primary prevention of stroke: a statement for healthcare professionals
  from the american heart association/american stroke association, Stroke
  45~(12) (2014) 3754--3832.

\bibitem{harmsen2006long}
P.~Harmsen, G.~Lappas, A.~Rosengren, L.~Wilhelmsen, Long-term risk factors for
  stroke: twenty-eight years of follow-up of 7457 middle-aged men in goteborg,
  sweden, Stroke 37~(7) (2006) 1663--1667.

\bibitem{nwosu2019predicting}
C.~S. Nwosu, S.~Dev, P.~Bhardwaj, B.~Veeravalli, D.~John, Predicting stroke
  from electronic health records, in: 2019 41st Annual International Conference
  of the IEEE Engineering in Medicine and Biology Society (EMBC), IEEE, 2019,
  pp. 5704--5707.

\bibitem{pathan2020identifying}
M.~S. Pathan, Z.~Jianbiao, D.~John, A.~Nag, S.~Dev, Identifying stroke
  indicators using rough sets, IEEE Access 8 (2020) 210318--210327.

\bibitem{jeena2016stroke}
R.~S. {Jeena}, S.~{Kumar}, Stroke prediction using {SVM}, in: Proc.
  International Conference on Control, Instrumentation, Communication and
  Computational Technologies (ICCICCT), 2016, pp. 600--602.
\newblock \href {https://doi.org/10.1109/ICCICCT.2016.7988020}
  {\path{doi:10.1109/ICCICCT.2016.7988020}}.

\bibitem{hanifa2010stroke}
S.-M. Hanifa, K.~Raja-S, Stroke risk prediction through non-linear support
  vector classification models, International Journal of Advanced Research in
  Computer Science 1~(3) (2010).

\bibitem{luk2006does}
J.~K. Luk, R.~T. Cheung, S.~Ho, L.~Li, Does age predict outcome in stroke
  rehabilitation? {A} study of 878 {Chinese} subjects, Cerebrovascular Diseases
  21~(4) (2006) 229--234.

\bibitem{min2018development}
S.~N. Min, S.~J. Park, D.~J. Kim, M.~Subramaniyam, K.-S. Lee, Development of an
  algorithm for stroke prediction: a national health insurance database study
  in korea, European neurology 79~(3-4) (2018) 214--220.

\bibitem{singh2017stroke}
M.~S. Singh, P.~Choudhary, Stroke prediction using artificial intelligence, in:
  2017 8th Annual Industrial Automation and Electromechanical Engineering
  Conference (IEMECON), IEEE, 2017, pp. 158--161.

\bibitem{chantamitprediction}
P.~Chantamit-o, Prediction of stroke disease using deep learning model.

\bibitem{khosla2010integrated}
A.~Khosla, Y.~Cao, C.~C.-Y. Lin, H.-K. Chiu, J.~Hu, H.~Lee, An integrated
  machine learning approach to stroke prediction, in: Proceedings of the 16th
  ACM SIGKDD international conference on Knowledge discovery and data mining,
  2010, pp. 183--192.

\bibitem{hung2019development}
C.-Y. Hung, C.-H. Lin, T.-H. Lan, G.-S. Peng, C.-C. Lee, Development of an
  intelligent decision support system for ischemic stroke risk assessment in a
  population-based electronic health record database, PloS one 14~(3) (2019)
  e0213007.

\bibitem{teoh2018towards}
D.~Teoh, Towards stroke prediction using electronic health records, BMC medical
  informatics and decision making 18~(1) (2018) 1--11.
  


\bibitem{hung2017comparing}
C.-Y. Hung, W.-C. Chen, P.-T. Lai, C.-H. Lin, C.-C. Lee, Comparing deep neural
  network and other machine learning algorithms for stroke prediction in a
  large-scale population-based electronic medical claims database, in: 2017
  39th Annual International Conference of the IEEE Engineering in Medicine and
  Biology Society (EMBC), IEEE, 2017, pp. 3110--3113.

\bibitem{li2016integrated}
X.~Li, H.~Liu, X.~Du, P.~Zhang, G.~Hu, G.~Xie, S.~Guo, M.~Xu, X.~Xie,
  Integrated machine learning approaches for predicting ischemic stroke and
  thromboembolism in atrial fibrillation, in: AMIA Annual Symposium
  Proceedings, Vol. 2016, American Medical Informatics Association, 2016, p.
  799.

\bibitem{garcia2016tutorial}
S.~Garc{\'\i}a, J.~Luengo, F.~Herrera, Tutorial on practical tips of the most
  influential data preprocessing algorithms in data mining, Knowledge-Based
  Systems 98 (2016) 1--29.

\bibitem{goldstein2017opportunities}
B.~A. Goldstein, A.~M. Navar, M.~J. Pencina, J.~Ioannidis, Opportunities and
  challenges in developing risk prediction models with electronic health
  records data: a systematic review, Journal of the American Medical
  Informatics Association 24~(1) (2017) 198--208.

\bibitem{abdi2010principal}
H.~Abdi, L.~J. Williams, Principal component analysis, Wiley interdisciplinary
  reviews: computational statistics 2~(4) (2010) 433--459.

\bibitem{tilahun2014design}
B.~Tilahun, T.~Kauppinen, C.~Ke{\ss}ler, F.~Fritz, Design and development of a
  linked open data-based health information representation and visualization
  system: potentials and preliminary evaluation, JMIR medical informatics 2~(2)
  (2014).

\bibitem{orlandi2019interlinking}
F.~Orlandi, A.~Meehan, M.~Hossari, S.~Dev, D.~O'Sullivan, T.~AlSkaif,
  Interlinking heterogeneous data for smart energy systems, in: 2019
  International Conference on Smart Energy Systems and Technologies (SEST),
  IEEE, 2019, pp. 1--6.
  
\bibitem{wu2021uplifting}
J. Wu, F. Orlandi, I. Gollini, E. Pisoni, S. Dev,
  Uplifting Air Quality Data Using Knowledge Graph, in: 2021
  Photonics \& Electromagnetics Research Symposium (PIERS),
  IEEE, 2021, pp. 2347--2350.

\end{thebibliography}
\end{document}